\documentclass{article} 
\usepackage{iclr2026_conference,times}
\iclrfinalcopy
 
\usepackage{amsmath,amsfonts,bm}

\def\eqref#1{equation~\ref{#1}}

\def\1{\bm{1}}

\DeclareMathAlphabet{\mathsfit}{\encodingdefault}{\sfdefault}{m}{sl}
\SetMathAlphabet{\mathsfit}{bold}{\encodingdefault}{\sfdefault}{bx}{n}

\usepackage{hyperref}
\usepackage{cleveref}
\usepackage{algorithm, algorithmic}
\usepackage{url}
\usepackage{multirow}
\usepackage{tikz}
\usepackage{booktabs}
\usepackage{subcaption}
\usetikzlibrary{positioning, shapes, arrows.meta, calc, shadows.blur}

\title{CircuitBuilder: From Polynomials to Circuits
\\ via Reinforcement Learning}

\author{
Weikun K. Zhang\thanks{Equal contribution.} \ \thanks{University of Washington}
\And
Rohan Pandey\footnotemark[1] \  \footnotemark[2]
\And
Bhaumik Mehta\footnotemark[2]
\And
Kaijie Jin\footnotemark[2]
\And
Naomi Morato\footnotemark[2]
\And
Archit Ganapule\footnotemark[2]
\And
Michael Ruofan Zeng\footnotemark[2] \  \thanks{Correspondence to: Michael R. Zeng <zengrf@uw.edu>}
\And
Jarod Alper\footnotemark[2]
}

\begin{document}

\maketitle
\fancyhead{}
\renewcommand{\headrulewidth}{0pt}

\begin{abstract}
Motivated by auto-proof generation and Valiant's VP vs. VNP conjecture, we study the problem of discovering efficient arithmetic circuits to compute polynomials, using addition and multiplication gates. We formulate this problem as a single-player game, where an RL agent attempts to build the circuit within a fixed number of operations. We implement an AlphaZero-style training loop and compare two approaches: Proximal Policy Optimization with Monte Carlo Tree Search (PPO+MCTS) and Soft Actor-Critic (SAC). SAC achieves the highest success rates on two-variable targets, while PPO+MCTS scales to three variables and demonstrates steady improvement on harder instances. These results suggest that polynomial circuit synthesis is a compact, verifiable setting for studying self-improving search policies.

\end{abstract}


\section{Introduction}

The search for the most efficient way to compute a polynomial is a foundational problem in algebraic complexity theory (see for example \citet{Burgisser1997AlgebraicComplexity, ShpilkaYehudayoff2009Circuits}). Known as the \textit{arithmetic circuit problem}, this task involves finding a sequence of addition and multiplication gates that compute a target polynomial $f(x_{1}, \dots, x_{n})$ using the minimum number of operations. This problem is more than a theoretical exercise. It is the algebraic analogue of the P vs. NP question, where the classes 
\begin{itemize}
    \item VP -- families of polynomials computable by polynomial-size circuits, and
    \item VNP -- families of polynomials whose coefficients are computable in polynomial time, which is the algebraic analogue of NP,
\end{itemize} represent the limits of efficient computation \citep{Valiant1979CompletenessClasses}. The understanding of the circuit size of certain families of polynomials such as the permanents could lead to huge progress in the VP vs. VNP problem \citep{Valiant1979Permanent}.

Furthermore, finding minimal circuits is a simpler version of the search for short mathematical proofs, where a sequence of logical deductions leads from axioms to a theorem. The arithmetic circuit problem is an ideal testing ground for which RL approaches might succeed in auto-proof generation. By systematically studying the effectiveness of various algorithms to search for arithmetic circuits, we hope to gain insight into the problem of proof search \citep{Hubert2025AlphaProof}.

However, the search space for such circuits is vast. For a circuit with $k$ intermediate nodes, the number of possible next operations grows at a rate of $O(k^{2})$, leading to an exponential growth that renders exhaustive search impractical for complex polynomials. Historically, efficient constructions such as the Horner scheme for univariate polynomials or recursive structures for elementary symmetric polynomials have been discovered through human intuition (see \Cref{subsec:circuits}). In this work, we investigate whether machine learning agents can autonomously discover these and other highly efficient computational structures.

Inspired by the success of AlphaZero \citep{Silveretal2018AlphaZero} in mastering complex two-player games and AlphaProof \citep{Hubert2025AlphaProof} in auto-proof generation, we model arithmetic circuit construction as a single-player Markov Decision Process (MDP). In this environment, an agent starts with input variables and constants and must select a sequence of algebraic operations to reach a target polynomial. Our primary metric for success is the agent's ability to accurately compute the given polynomial by reaching the most efficient target - a minimal gate count - while successfully generalizing to unseen polynomials.

Our contribution is a comparative study of two distinct architectural approaches to this problem:
\begin{itemize}
    \item \underline{CircuitBuilder (PPO + MCTS):} A reinforcement learning agent that uses Proximal Policy Optimization (PPO) combined with Monte Carlo Tree Search (MCTS) to guide exploration through the massive search space and sparse reward signals.
    \item \underline{Soft Actor-Critic (SAC):} An off-policy actor-critic method to address the challenges of sparse reward signals, which frequently hinder training progression in circuit discovery tasks \citep{Haarnoja2018SAC}.
\end{itemize}

We evaluate these methods on previously unseen polynomials, where we demonstrate that deep search and structured learning can recover optimal or near-optimal circuits. We also examine strategies to efficiently sample polynomials with multiple optimal circuits, to improve training and testing. Our results suggest that learning-based approaches can provide new insights into algebraic complexity and offer a scalable path toward a recursively self-improving agent for mathematical discoveries. 

\section{Background}

\subsection{Arithmetic circuits}
\label{subsec:circuits}

We summarize some facts about arithmetic circuits following \citet{Burgisser1997AlgebraicComplexity}. Let $\mathbb F$ be a field or more generally a commutative ring. Let $X=\{x_1,\dots,x_n\}$ be formal variables. An \textit{arithmetic circuit} with coefficients in $\mathbb F$ and variables $X$ is a finite directed acyclic graph (DAG) whose vertices (called ``gates'') have indegree $0$ or $2$. Gates of indegree $0$ are \textit{input gates} labeled by variables $X$. Gates of indegree $2$ are \textit{addition / multiplication gates} labeled by $+$ or $\times$. One distinguished gate is designated as the \textit{output gate}. Note that \citet{Burgisser1997AlgebraicComplexity} allows input gates valued in possibly negative constants. We stay within the scope of variable-only input gates (one of the variables can be viewed as the constant `$1$') and defer other constant values to future work.

\begin{figure}
    \centering
   \begin{tikzpicture}[
    scale=0.5,
    op/.style={circle, draw, fill=orange!10, minimum size=0.4cm, thick, font=\large},
    var/.style={rectangle, draw, fill=blue!10, minimum size=0.5cm, thick, font=\large},
    edge/.style={->, >=stealth, thick}
]

\begin{scope}[rotate=-90, transform shape=false]
    \node[var] (x1) at (-5,0) {$x$};
    \node[var] (x2) at (-2.5,0) {$y$};
    \node[op] (add) at (-3.75,3) {$+$};
    \node[op] (mult) at (-3.75,6) {$\times$};
    
    \draw[edge] (x1) -- (add);
    \draw[edge] (x2) -- (add);
    \draw[edge] (add) to[bend left=15] (mult);
    \draw[edge] (add) to[bend right=15] (mult);

    \node[below=1 cm of add] {(Efficient)};
    \node[above=1 cm of add, font=\large] {$(x+y)^2$};
\end{scope}

\draw[dashed, gray] (8, 1) -- (8, 7);

\begin{scope}[shift={(12,0)}, rotate=-90, transform shape=false]
    \node[var] (rx) at (-5,0) {$x$};
    \node[var] (ry) at (-2.5,0) {$y$};

    \node[op] (mx2) at (-5.5,3) {$\times$};
    \node[op] (mxy) at (-3.75,3) {$\times$};
    \node[op] (my2) at (-2,3) {$\times$};

    \node[op] (radd1) at (-3.75,6) {$+$};
    \node[op] (radd2) at (-4.625,9) {$+$};
    \node[op] (radd3) at (-2.875,12) {$+$};

    \draw[edge] (rx) -- (mx2);
    \draw[edge] (rx) to[bend left=20] (mx2);

    \draw[edge] (rx) -- (mxy);
    \draw[edge] (ry) -- (mxy);

    \draw[edge] (ry) -- (my2);
    \draw[edge] (ry) to[bend right=20] (my2);

    \draw[edge] (mxy) to[bend left=15] (radd1);
    \draw[edge] (mxy) to[bend right=15] (radd1);

    \draw[edge] (mx2) to[bend left=15] (radd2);
    \draw[edge] (radd1) to[bend right=15] (radd2);

    \draw[edge] (radd2) to[bend left=10] (radd3);
    \draw[edge] (my2) to[bend right=15] (radd3);

    \node[below=1 cm of radd1] {(Inefficient)};
    \node[above=1 cm of radd1, font=\large] {$x^2 + 2xy + y^2$};
\end{scope}

\end{tikzpicture}
    \caption{Two arithmetic circuits for the polynomial $x^2 + 2xy+y^2$.}
    \label{fig:circuits}
\end{figure}
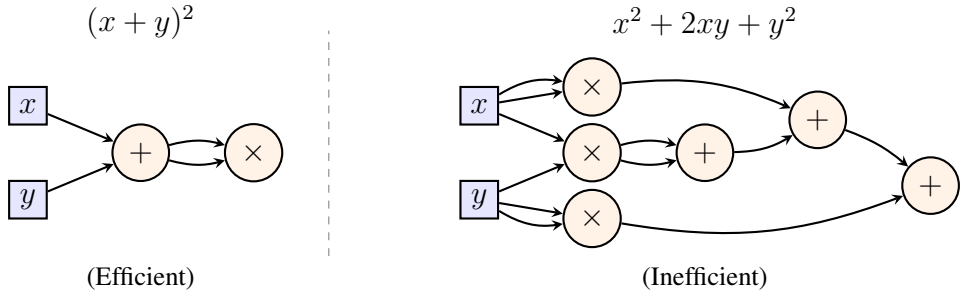

Each gate $v$ computes a polynomial $f_v\in \mathbb F[X]$ defined inductively. The input gates represent the monomials $x_1, \dots, x_n$. If $v$ is a $+$ / $\times$-gate with incoming neighbors $u,w$, then $f_v=f_u+f_w$ (respectively\ $f_v=f_u\cdot f_w$). The circuit \textit{computes} a polynomial $f$ if the polynomial at the output gate equals $f$. The \textit{complexity} of a circuit is the number of gates, and the \textit{depth} is the length of the longest directed path from an input to the output. The \textit{syntactic degree} is defined inductively by $\deg(v)=0$ for constant inputs, $\deg(v)=1$ for variable inputs, $\deg(v)=\max\{\deg(u),\deg(w)\}$ at $+$-gates, and $\deg(v)=\deg(u)+\deg(w)$ at $\times$-gates. It upper-bounds the total degree of $f_v$. See \Cref{fig:circuits} for two different arithmetic circuits for the polynomial $x^2 + 2xy + y^2$. 



Many polynomials admit small arithmetic circuits because they can be computed by reusing intermediate subexpressions. A first example is the \textit{Horner scheme} \citep{Horner1819NewMethod}. Given a univariate polynomial
\(
p(x)=\sum_{i=0}^{d} a_i x^i,
\)
one may evaluate $p$ by the decreasing recurrence
\(
t_d=a_d,\qquad t_i=a_i + x\cdot t_{i+1}\quad (i=d-1,\dots,0),
\)
so that $p(x)=t_0$. This recursion is equivalent to the nested expression
\[
p(x)=a_0 + x\bigl(a_1 + x( a_2 + \cdots + x(a_{d-1}+x a_d)\cdots )\bigr).
\]
Horner's method uses exactly $d$ addition and $d$ multiplication gates, in contrast to the naive circuit $\sum_i a_i x^i$ which recomputes powers $x^i$ for each term. 

A multivariate example is provided by \textit{elementary symmetric polynomials}. The elementary symmetric polynomial of degree $k$ in $n$ variables is
\(
e_k(x_1,\dots,x_n)=\sum_{1\le i_1<\cdots<i_k\le n} x_{i_1}\cdots x_{i_k},
\)
and it satisfies the recurrence
\(
e_k(x_1,\dots,x_n)=e_k(x_1,\dots,x_{n-1}) + x_n\cdot e_{k-1}(x_1,\dots,x_{n-1}).
\)
This recurrence expresses $e_k$ in terms of two previously computed polynomials on $(n-1)$ variables.

The arithmetic circuit problem is central in algebraic complexity theory. The class $\mathrm{VP}$ consists of families of polynomials computable by arithmetic circuits of polynomial size in terms of the parameter \citep{Valiant1979CompletenessClasses}. Valiant's $\mathrm{VNP}$ class is an algebraic analogue of $\mathrm{NP}$. Valiant then showed that computing the \textit{permanent polynomials}

\[\operatorname{per}_n\left(x_{i,j}\right)=\sum_{\sigma \in S_n} \prod_{i=1}^n x_{i, \sigma(i)}, \quad  (i,j) \in [n] \times [n]\]
is $\mathrm{VNP}$-complete \citep{Valiant1979Permanent}.

\subsection{AlphaZero-Style Search and Policy Optimization}

AlphaZero's achievement was its ability to master two-player perfect information games (such as Go, chess, shogi) \citep{Silveretal2018AlphaZero}. AlphaZero combined a neural network with the MCTS framework, allowing the agent to master gameplay through self-play only without any human feedback/input beyond the rules of the game. Central to the approach is a single neural network that outputs both a policy (probability distribution over moves to choose) and a value estimate (predicted outcome from a given state), which together guide the tree search.

\paragraph{\underline{Monte Carlo Tree Search}}

The Monte-Carlo Tree Search algorithm is a powerful search algorithm that samples value-based trajectories to incrementally build a search tree. It balances exploration and exploitation using upper confidence bounds (UCB)  \citep{KocsisBanditMonteCarlo}. For a state-action pair $(s,a)$, one has $$\text{UCB}(s, a) = \frac{Q(s, a)}{N(s, a)} + c \sqrt{\frac{\ln N(s)}{N(s, a)}},$$

where $Q(s, a)$ is the total value, $N(s, a)$ is the visit count for that action, $N(s)$ is the parent visit count, and $c$ is an exploration constant that balances exploitation of high-value actions with exploration of less-visited nodes.

The algorithm proceeds in four phases: selection, where the tree is traversed using UCB; expansion, where a new node is added; rollout, where the value of the new node is estimated; and backpropagation, where statistics are updated along the visited path.

The sequential nature of circuit construction (selecting one gate at a time) maps naturally onto MCTS's tree structure, making it well-suited for navigating the combinatorial search space of arithmetic circuits.

\paragraph{\underline{Proximal Policy Optimization}}

Proximal Policy Optimization (PPO) is a policy-gradient reinforcement learning method; its key idea is that it stabilizes training by preventing the policy from changing drastically in a single update step. The core equation of PPO is a clipped surrogate function \citep{schulman2017ppo} $$L^{\text{CLIP}}(\theta) = \mathbb{E}\left[ \min\left(r_t(\theta)\hat{A}_t, \text{clip}(r_t(\theta), 1 - \epsilon, 1 + \epsilon)\hat{A}_t\right) \right],$$

where $r_t(\theta) = \frac{\pi_\theta(a_t|s_t)}{\pi_{\theta_{\text{old}}}(a_t|s_t)}$ is the probability ratio between the new and old policies, $\hat{A}_t$ is the estimated advantage and $\epsilon$ is a clipping hyperparameter which helps prevent updates that are unhelpfully large.

In our work, we implemented PPO as the training algorithm for the neural network. This replaces AlphaZero's original implementation of the policy iteration scheme. PPO's training stability is particularly valuable in our setting, where the reward landscape of circuit construction is sparse and sensitive to large policy shifts.

Together, the neural network trained via PPO provides the policy prior and value estimates that guide MCTS, while the improved action distributions produced by MCTS serve as training targets for the network to choose the best action.

\paragraph{\underline{Soft Actor-Critic}}

SAC is an off-policy actor-critic method that adds the standard RL objective function with an entropy bonus which encourages the agent to continue exploration. SAC employs twin Q-networks to reduce overestimation bias, along with a learnable temperature parameter $\alpha$ that balances how much the agent prioritizes rewards vs. exploration \citep{Haarnoja2018SAC}. The SAC objective function is $$J(\pi) = \sum_t \mathbb{E}\left[ r(s_t, a_t) + \alpha H(\pi(\cdot | s_t)) \right],$$ where $H$ is the entropy of the policy. The agent's goal is to maximize the reward while also keeping its policy stochastic. In our work, the circuit construction has sparse rewards, so the agent might not know if it's on the right track until many steps are taken, and SAC's entropy regularization helps maintain exploration in this setting. While SAC was originally designed for continuous action spaces, we adapt it to discrete settings \citep{christodoulou2019sacdiscrete}.



\section{Methodology}
\label{sec:methods}
\subsection{Environment and State Encoding}

We model arithmetic circuit construction as a single-player Markov Decision Process (MDP), where the state is a partially-built circuit represented as a directed acyclic graph (DAG), actions correspond to selecting an operation and two existing nodes, and transitions deterministically append a new node to the graph. The agent's goal is to reach a state where one of the computed nodes matches the target polynomial, in as few steps as possible.

Formally, let $\mathcal{G}_t = (V_t, E_t)$ denote the circuit graph at step $t$, with initial nodes $V_0 = \{x_0, \ldots, x_{n-1}, 1\}$ consisting of the input variables and a constant. At each step the agent selects an action $a_t = (\star, v_i, v_j)$ where $\star \in \{+, \times\}$ and $v_i, v_j \in V_t$, producing a new node $v_{\text{new}}$ that computes $v_i \star v_j$. The transition updates the graph as $V_{t+1} = V_t \cup \{v_{\text{new}}\}$.

Each node $v \in V_t$ is represented by a 4-dimensional feature vector consisting of a 3-bit one-hot encoding for node type (input variable, constant, or operation result) and a scalar value, with directed edges connecting operand nodes to their result nodes and self-loops added for message passing. The target polynomial is not represented symbolically but as a compact circuit encoding. This encoding concatenates operation-type one-hots, edge-selection one-hots, and a last-generated-node indicator derived from a reference action sequence. The action space is also flattened, so each action $a_t = (\star, v_i, v_j)$ is mapped to a unique integer index. Invalid actions (referencing nodes that are not yet created) are masked out at each step.

\subsection{Game-Board Generation}


To generate structured training data, we construct a ``game-board'' directed acyclic graph (DAG) enumerating all polynomials reachable from seed nodes $V_0 = \{x_0, \ldots, x_{n-1}, 1\}$ within $\mathcal{C}$ arithmetic operations. At each step, pairs of existing nodes are combined via addition and multiplication. A node's appearance step corresponds to the minimum number of gates needed to compute that polynomial. In this graph, a circuit corresponds to a path from a root node in $V_0$ to a target polynomial, and an \textit{optimal circuit} is defined as a shortest such path.

Not all polynomials in the DAG are equally useful for training. We define a polynomial as \textit{interesting} if it admits multiple distinct shortest paths in the DAG (i.e., multiple optimal circuits). Targeting these polynomials forces the agent to learn meaningful decision-making rather than memorizing a single forced sequence.



Table \ref{tab:c4_depth} reports the distribution of nodes by circuit depth. Most nodes appear at depths three and four, reflecting the rapid combinatorial growth of reachable polynomials. Table \ref{tab:c4_summary} summarizes the structural properties of these graphs. In both the two-variable (C4-main) and single-variable (C4-pretrain) boards, the vast majority of nodes admit multiple optimal circuits, confirming the environment contains substantial decision freedom. Ultimately, the search space grows exponentially in $C$. The number of different circuits with $n$ variables and complexity $\le C$ over $\mathbb{F}_p$ is bounded by $\exp (\Theta(C \log (n+p+C)))$.

\begin{table}[t]
\centering
\small
\begin{tabular}{ccc}
\toprule
Depth  & C4-main & C4-pretrain \\
\midrule
0 & 3 & 1 \\
1 & 12 & 2 \\
2 & 174 & 9 \\
3 & 10{,}862 & 96 \\
4 & 8{,}949 & 6{,}548 \\
\bottomrule
\end{tabular}
\caption{Circuit-depth distribution for C4 boards.}
\label{tab:c4_depth}
\end{table}

\begin{table}[t]
\centering
\small
\begin{tabular}{lcc}
\toprule
Metric & C4-main (multivar) & C4-pretrain (single-var) \\
\midrule
Nodes & 20{,}000 & 6{,}656 \\
Edges & 31{,}746 & 20{,}966 \\
Roots & 3 & 1 \\
Nodes with multiple optimal circuits & 18{,}966 (94.83\%) & 6{,}592 (99.04\%) \\
Max optimal circuits for one node & 164 & 92 \\
Max total circuits for one node & 1{,}596 & 15{,}296 \\
\bottomrule
\end{tabular}
\caption{C4 game-board structure and path-efficiency statistics. Optimal circuits are shortest root-to-node paths.}
\label{tab:c4_summary}
\end{table}

\subsection{PPO + MCTS}

Our primary agent, CircuitBuilder, combines graph-based state encoding with a Transformer decoder to produce both a policy and value estimate. The GNN encoder processes the circuit DAG. It uses GCNConv layers with residual connections and layer normalization, and the per-node embeddings are aggregated via a global mean pooling into a single graph embedding vector. The target polynomial is encoded using a compact one-hot scheme that concatenates operation-type indicators, edge-selection indicators, and a last-generated-node indicator from a reference action sequence. This is then projected through a linear layer into the same embedding dimension. A Transformer decoder then attends over both embeddings: a learnable output token serves as the query, while the graph embedding and target polynomial embedding are stacked to form the decoder memory. The decoder output is passed to a policy head, which produces logits over the action space masked to valid actions, and a value head, which outputs a scalar estimate of the current state's value.

\paragraph{\underline{Supervised Pretraining}}

We first train on (state, next-action) pairs extracted from known optimal circuits from the game board, where each intermediate step of a circuit becomes a training example. The model is trained to minimize cross-entropy loss on action predictions plus mean squared error on value predictions. This creates a strong policy initialization before RL fine-tuning.

\paragraph{\underline{PPO Fine-Tuning}}

After supervised pretraining, we fine-tune the model using PPO. We use Generalized Advantage Estimation (GAE) to compute low-variance advantage estimates. $$\delta_t = r_t + \gamma V(s_{t+1}) - V(s_t), \qquad \hat{A}_t = \delta_t + \gamma \lambda \hat{A}_{t+1}.$$
Here, $\gamma$ is the discount factor and $\lambda$ controls the bias-variance tradeoff. The full PPO loss function that we optimize is $$\mathcal{L}_{\text{PPO}}(\theta) = -\mathbb{E}\left[\min\left(r_t(\theta)\hat{A}_t, \text{clip}(r_t(\theta), 1-\epsilon, 1+\epsilon)\hat{A}_t\right)\right] + c_v \text{MSE}(V_\theta(s_t), \hat{R}_t) - c_e \mathbb{E}[\mathcal{H}(\pi_\theta(\cdot|s_t))].$$

The three terms are the clipped surrogate objective, a value function loss, and an entropy bonus that encourages exploration. We employ a curriculum learning strategy: training begins at polynomial complexity 1, and the complexity increases when the agent's success rate exceeds a threshold over a sliding window.

\paragraph{\underline{MCTS Integration}}

During PPO data collection, MCTS optionally guides action selection. We employ the Bernoulli mixing scheme: $$a_t = \begin{cases} a_t^{\text{MCTS}}, & z_t = 1 \\ a_t^{\pi}, & z_t = 0 \end{cases}, \quad z_t \sim \text{Bernoulli}(p_{\text{mix}}).$$

With probability $p_\text{mix}$ the agent defers to the MCTS planner, otherwise it samples from its own policy. Our MCTS uses the neural network's value head to evaluate leaf nodes (AlphaZero-style), replacing random rollouts with learned value estimates. It returns the action with the highest visit count. This exposes the policy network to higher-quality trajectories during training, improving sample efficiency over pure policy sampling.
\subsubsection{MCTS-Guided Expert Iteration}

At each timestep $t$, the MCTS search is guided by the current policy-value network, producing a visit-count distribution $\pi_\text{MCTS}$. To balance early-episode exploration with later-episode exploitation, action selection is tempered by a decaying temperature parameter $\tau$:
$$
\tau(t) = \tau_\text{final} + (\tau_\text{init} - \tau_\text{final}) \max \bigl(1 - \frac{t}{t_\text{decay}}, 0 \bigr),
$$
where $t_\text{decay}$ defines the annealing schedule.

A critical architectural distinction in our approach lies in the formulation of the PPO importance sampling ratio. Standard AlphaZero implementations train the policy head to minimize the cross-entropy loss against the MCTS visit counts. In our PPO formulation, the surrogate objective relies on the probability ratio
$$
r_t(\theta) = \frac{\pi_{\theta}(a_t|s_t)}{\pi_{\theta_\text{old}}(a_t|s_t)}.
$$
Crucially, the denominator $\pi_{\theta_\text{old}}(a_t|s_t)$ is the probability assigned by the \textit{network's own policy} at the time of data collection, not the probability derived from the MCTS visit counts. If the MCTS probabilities were used in the denominator, the ratio $r_t(\theta)$ would remain near $1$, resulting in a near-zero policy gradient. 

By using the network's internal logits for the baseline ratio, MCTS acts strictly as a data-quality enhancer—finding shorter, more efficient circuit paths—while the clipped PPO objective provides stable, meaningful updates to the network parameters to approximate this improved behavior. Furthermore, the Generalized Advantage Estimation (GAE) targets are bootstrapped using the network's value head $V_\theta(s)$, rather than the MCTS value estimates, ensuring consistent variance reduction during the PPO update.


\subsection{Soft Actor-Critic}

The SAC method shares the same GNN-Transformer backbone as CircuitBuilder, but it replaces the single value head with twin Q-heads that output per-action Q-values. The SAC is off-policy, storing transitions in a replay buffer and using soft target network updates via Polyak averaging. See Algorithm~\ref{alg:sac} \citep{OpenAISAC} in Appendix \ref{apx:algorithms} for the full update procedure. 

We employ the same curriculum learning strategy as PPO, with the addition that complexity can also decrease when the success rate falls below a lower threshold, preventing the agent from stalling on targets beyond its current capability.




\section{Results}
\label{sec:results}

We trained PPO+MCTS and SAC on an AWS EC2 cloud machine with an NVIDIA Tesla T4 GPU (16\,GB). Both models use a GNN encoder with \texttt{GCNConv} layers to process the circuit DAG, fused with a target polynomial embedding, and output policy logits and value estimates. We evaluate on fixed-complexity targets at $C=5$ and $C=6$ over $\mathbb{F}_5$ for both two and three variables, using 1000 held-out episodes per checkpoint with near-greedy MCTS ($\tau=0.1$).

For a target polynomial $f$, we declare \emph{success} if the agent produces an arithmetic circuit whose output polynomial equals $f$. Table~\ref{tab:ppo_summary} reports evaluation metrics for PPO+MCTS over 200 training iterations with batch size 256.

\begin{table}[h]
\centering
\caption{PPO+MCTS evaluation on $\mathbb{F}_5$ (1000 episodes per checkpoint, near-greedy MCTS with $\tau=0.1$). Success rates are percentages; entropy is the policy entropy at iter 200.}
\label{tab:ppo_summary}
\begin{tabular}{ll c c c c c}
\toprule
Variables & Task & Iter 50 & Iter 100 & Iter 150 & Iter 200 & Entropy \\
\midrule
\multirow{4}{*}{$n=2$}
& $C=5$ success (\%) & 29.40 & 26.50 & 27.40 & \textbf{34.90} & 3.28 \\
& $C=6$ success (\%) & 26.00 & 27.30 & 27.60 & \textbf{35.80} & 3.21 \\
& $C=5$ avg reward   & 2.47 & 2.22 & 2.34 & 3.08 & --- \\
& $C=6$ avg reward   & 2.19 & 2.30 & 2.31 & 3.20 & --- \\
\midrule
\multirow{4}{*}{$n=3$}
& $C=5$ success (\%) & 17.70 & 20.90 & 26.50 & \textbf{27.30} & 3.65 \\
& $C=6$ success (\%) & 16.40 & \textbf{23.50} & 20.90 & 19.90 & 3.72 \\
& $C=5$ avg reward   & 1.34 & 1.58 & 2.21 & 2.30 & --- \\
& $C=6$ avg reward   & 1.18 & 1.93 & 1.67 & 1.53 & --- \\
\bottomrule
\end{tabular}
\end{table}

For $n=2$ variables, PPO+MCTS reaches ${\sim}35\%$ success on both $C=5$ and $C=6$ by iteration 200, with entropy decreasing slightly from ${\sim}3.3$ to ${\sim}3.2$, indicating policy sharpening without collapse.

Scaling to $n=3$ is notably harder: the target space grows from $7^2=49$ to $7^3=343$ coefficients. At $C=5$ the agent still improves monotonically to 27.3\%, but $C=6$ peaks at iteration 100 (23.5\%) then degrades, suggesting training instability in the larger search space. The higher entropy for $n=3$ (${\sim}3.7$ vs.\ ${\sim}3.2$) reflects greater uncertainty over the expanded action space.

\begin{table}[h]
\centering
\caption{SAC evaluation on $\mathbb{F}_5$ (1000 episodes per checkpoint, greedy policy). Success rates are percentages; entropy is the policy entropy at iter 200.}
\label{tab:sac_summary}
\begin{tabular}{ll c c c c c}
\toprule
Variables & Task & Iter 50 & Iter 100 & Iter 150 & Iter 200 & Entropy \\
\midrule
\multirow{4}{*}{$n=2$}
& $C=5$ success (\%) & 48.30 & 55.60 & 53.20 & \textbf{57.80} & 1.85 \\
& $C=6$ success (\%) & 45.10 & \textbf{49.70} & 48.30 & 46.30 & 1.88 \\
& $C=5$ avg reward   & 3.41 & 4.14 & 3.90 & 4.37 & --- \\
& $C=6$ avg reward   & 3.09 & 3.52 & 3.39 & 3.19 & --- \\
\midrule
\multirow{4}{*}{$n=3$}
& $C=5$ success (\%) & 5.40 & 5.00 & 6.20 & \textbf{10.70} & 2.23 \\
& $C=6$ success (\%) & 4.30 & 4.30 & 5.60 & \textbf{9.40} & 2.22 \\
& $C=5$ avg reward   & $-$0.99 & $-$1.03 & $-$0.89 & $-$0.45 & --- \\
& $C=6$ avg reward   & $-$1.10 & $-$1.09 & $-$0.95 & $-$0.58 & --- \\
\bottomrule
\end{tabular}
\end{table}

Table~\ref{tab:sac_summary} reports evaluation metrics for SAC over the same 200 training iterations with batch size 256. For \(n=2\), SAC performs reasonably well and reaches its best results at iter 200, with \(57.8\%\) success on \(C=5\) targets and \(46.3\%\) on \(C=6\). The average rewards are also positive throughout and generally improve over training, suggesting that SAC is able to learn useful symbolic construction strategies in the two-variable setting. As expected, performance on \(C=6\) remains below \(C=5\), reflecting the greater difficulty of searching over more complex targets.

For \(n=3\), however, performance drops sharply. Even at iter 200, SAC achieves only \(10.7\%\) success on \(C=5\) and \(9.4\%\) on \(C=6\), while the average rewards remain negative despite some improvement. Compared with the relatively small gap between \(C=5\) and \(C=6\), the much larger drop from \(n=2\) to \(n=3\) suggests that increasing the number of variables is a more significant source of difficulty. This indicates that future improvements should focus on better exploration, stronger representations, and training strategies that scale more effectively to higher-dimensional polynomial spaces.

\paragraph{\underline{Discussion \& Future Directions}}
Based on the updated results, SAC achieves substantially stronger performance than PPO+MCTS in the two-variable setting.
This suggests that, in the present setup, SAC is more effective at translating training experience into successful symbolic construction policies. Its average reward also improves over training, indicating steady learning progress, although the decline in entropy suggests that exploration becomes more concentrated as training proceeds. Notably, PPO+MCTS with learned tree search scales more gracefully to three variables than SAC, suggesting that explicit planning via MCTS may become increasingly important as problem dimensionality grows.

The remaining gap between \(C=5\) and \(C=6\) still reflects the rapid growth of the symbolic search space as circuit complexity increases. Even a single additional operation greatly enlarges the number of reachable expressions, making it harder for the agent to consistently identify successful constructions within a fixed horizon. Future work includes extending training to higher-complexity targets (\(C \geq 7\)), investigating stronger exploration strategies for harder instances, and incorporating curriculum learning across multiple complexity levels. Another important direction is to improve polynomial and circuit representations so that the agent can generalize more effectively as the search space grows.

Our experiments still focus on relatively small circuits, with three variables and target complexities up to six gates. Although this setting is far smaller than the regimes studied in algebraic complexity theory, it provides a controlled environment for testing whether reinforcement learning can learn meaningful circuit-construction strategies. The purpose of this stage is to understand learning dynamics on small symbolic problems before scaling to larger circuits, more variables, and richer polynomial families in future work.

\subsubsection*{\underline{Acknowledgments}}
This project is a part of the UW Math AI Lab. We thank the UW eScience School for computing resources. CPU and GPU computing were in part done using AWS credits from the UW eScience School and UW IT, and also in part done using the UW Research Computing Club funded from the UW Student Technology Fee Committee. Some parts of the code base are produced with the help of GitHub Copilot.

\newpage

\bibliography{iclr2026_conference}

@article{ShpilkaYehudayoff2009Circuits,
 author = {Shpilka, Amir and Yehudayoff, Amir},
 title = {Arithmetic circuits: a survey of recent results and open questions},
 fjournal = {Foundations and Trends in Theoretical Computer Science},
 journal = {Found. Trends Theor. Comput. Sci.},
 issn = {1551-305X},
 volume = {5},
 number = {3-4},
 pages = {207--388},
 year = {2009},
 language = {English},
 doi = {10.1561/0400000039},
 zbMATH = {5840661},
 Zbl = {1205.68175}
}

@article{Silveretal2018AlphaZero,
	title = {A general reinforcement learning algorithm that masters chess, shogi, and {Go} through self-play},
	volume = {362},
	issn = {0036-8075, 1095-9203},
	url = {https://www.science.org/doi/10.1126/science.aar6404},
	doi = {10.1126/science.aar6404},
	language = {en},
	number = {6419},
	journal = {Science},
	author = {Silver, David and Hubert, Thomas and Schrittwieser, Julian and Antonoglou, Ioannis and Lai, Matthew and Guez, Arthur and Lanctot, Marc and Sifre, Laurent and Kumaran, Dharshan and Graepel, Thore and Lillicrap, Timothy and Simonyan, Karen and Hassabis, Demis},
	month = dec,
	year = {2018},
	pages = {1140--1144},
}

@article{schulman2017ppo,
  title={Proximal Policy Optimization Algorithms},
  author={Schulman, John and Wolski, Filip and Dhariwal, Prafulla and Radford, Alec and Klimov, Oleg},
  journal={arXiv preprint arXiv:1707.06347},
  year={2017}
}

@article{Hubert2025AlphaProof,
  title   = {Olympiad-level formal mathematical reasoning with reinforcement learning},
  author  = {Hubert, Thomas and Mehta, Rishi and Sartran, Laurent and Horváth, Miklós Z. and Žužić, Goran and Wieser, Eric and Huang, Aja and Schrittwieser, Julian and Schroecker, Yannick and Masoom, Hussain and Bertolli, Ottavia and Zahavy, Tom and Mandhane, Amol and Yung, Jessica and Beloshapka, Iuliya and Ibarz, Borja and Veeriah, Vivek and Yu, Lei and Nash, Oliver and Lezeau, Paul and Mercuri, Salvatore and Sönne, Calle and Mehta, Bhavik and Davies, Alex and Zheng, Daniel and Pedregosa, Fabian and Li, Yin and von Glehn, Ingrid and Rowland, Mark and Albanie, Samuel and Velingker, Ameya and Schmitt, Simon and Lockhart, Edward and Hughes, Edward and Michalewski, Henryk and Sonnerat, Nicolas and Hassabis, Demis and Kohli, Pushmeet and Silver, David},
  journal = {Nature},
  year    = {2025},
  month   = {nov},
  day     = {12},
  doi     = {10.1038/s41586-025-09833-y},
  url     = {https://doi.org/10.1038/s41586-025-09833-y},
  issn    = {1476-4687}
}

@misc{Haarnoja2018SAC,
      title={Soft Actor-Critic: Off-Policy Maximum Entropy Deep Reinforcement Learning with a Stochastic Actor}, 
      author={Tuomas Haarnoja and Aurick Zhou and Pieter Abbeel and Sergey Levine},
      year={2018},
      eprint={1801.01290},
      archivePrefix={arXiv},
      primaryClass={cs.LG},
      url={https://arxiv.org/abs/1801.01290}, 
}

@book{Burgisser1997AlgebraicComplexity,
 author = {B{\"u}rgisser, Peter and Clausen, Michael and Shokrollahi, M. Amin},
 title = {Algebraic complexity theory. {With} the collaboration of {Thomas} {Lickteig}},
 fseries = {Grundlehren der Mathematischen Wissenschaften},
 series = {Grundlehren Math. Wiss.},
 issn = {0072-7830},
 volume = {315},
 isbn = {3-540-60582-7},
 year = {1997},
 publisher = {Berlin: Springer},
 language = {English},
 zbMATH = {976329},
 Zbl = {1087.68568}
}

@InProceedings{KocsisBanditMonteCarlo,
author="Kocsis, Levente
and Szepesv{\'a}ri, Csaba",
editor="F{\"u}rnkranz, Johannes
and Scheffer, Tobias
and Spiliopoulou, Myra",
title="Bandit Based Monte-Carlo Planning",
booktitle="Machine Learning: ECML 2006",
year="2006",
publisher="Springer Berlin Heidelberg",
address="Berlin, Heidelberg",
pages="282--293",
isbn="978-3-540-46056-5"
}

@misc{christodoulou2019sacdiscrete,
      title={Soft Actor-Critic for Discrete Action Settings}, 
      author={Petros Christodoulou},
      year={2019},
      eprint={1910.07207},
      archivePrefix={arXiv},
      primaryClass={cs.LG},
      url={https://arxiv.org/abs/1910.07207}, 
}

@article{Horner1819NewMethod,
    author = {Horner, William George},
    title = {XXI. A new method of solving numerical equations of all orders, by continuous approximation},
    journal = {Philosophical Transactions of the Royal Society of London},
    number = {109},
    pages = {308-335},
    year = {1819},
    month = {12},
    issn = {0261-0523},
    doi = {10.1098/rstl.1819.0023},
    url = {https://doi.org/10.1098/rstl.1819.0023},
    eprint = {https://royalsocietypublishing.org/rstl/article-pdf/doi/10.1098/rstl.1819.0023/1451612/rstl.1819.0023.pdf},
}

@inproceedings{Valiant1979CompletenessClasses,
author = {Valiant, L. G.},
title = {Completeness classes in algebra},
year = {1979},
isbn = {9781450374385},
publisher = {Association for Computing Machinery},
address = {New York, NY, USA},
url = {https://doi.org/10.1145/800135.804419},
doi = {10.1145/800135.804419},
booktitle = {Proceedings of the Eleventh Annual ACM Symposium on Theory of Computing},
pages = {249–261},
numpages = {13},
location = {Atlanta, Georgia, USA},
series = {STOC '79}
}

@article{Valiant1979Permanent,
title = {The complexity of computing the permanent},
journal = {Theoretical Computer Science},
volume = {8},
number = {2},
pages = {189-201},
year = {1979},
issn = {0304-3975},
doi = {https://doi.org/10.1016/0304-3975(79)90044-6},
url = {https://www.sciencedirect.com/science/article/pii/0304397579900446},
author = {L.G. Valiant}
}

@misc{OpenAISAC,
  author       = {{OpenAI}},
  title        = {Soft Actor-Critic ({SAC})},
  year         = {2018},
  url          = {https://spinningup.openai.com/en/latest/algorithms/sac.html},
  note         = {Accessed: 2026-02-09}
}
\bibliographystyle{iclr2026_conference}

\newpage

\appendix

\section{Algorithms}
\label{apx:algorithms}

\begin{algorithm}[h]
\caption{Discrete SAC}
\label{alg:sac}
\begin{algorithmic}[1]
\REQUIRE Replay buffer $\mathcal{B}$, policy network $\pi_\theta$, twin Q-networks $Q_{\theta,1}, Q_{\theta,2}$, target networks $Q_{\bar{\theta},1}, Q_{\bar{\theta},2}$, temperature $\alpha$, MCTS coefficient $\lambda_{\text{mcts}}$, soft update rate $\tau$
\FOR{each update step}
    \STATE Sample minibatch $\{(s_t, a_t, r_t, s_{t+1}, d_t, \pi^{\text{MCTS}}_t, m_t)\} \sim \mathcal{B}$
    \STATE \textit{// Compute soft value target over valid actions}
    \STATE $V_{\bar{\theta}}(s') \leftarrow \sum_{a \in \mathcal{A}_{\text{valid}}(s')} \pi_\theta(a|s')\left(\min_i Q_{\bar{\theta},i}(s',a) - \alpha \log \pi_\theta(a|s')\right)$
    \STATE $y_t \leftarrow r_t + \gamma(1 - d_t)\, V_{\bar{\theta}}(s_{t+1})$
    \STATE \textit{// Update twin Q-networks}
    \STATE $\mathcal{L}_Q \leftarrow \text{MSE}(Q_{\theta,1}(s_t, a_t),\, y_t) + \text{MSE}(Q_{\theta,2}(s_t, a_t),\, y_t)$
    \STATE \textit{// Update policy over valid actions}
    \STATE $\mathcal{L}_\pi \leftarrow \mathbb{E}_{s_t}\!\left[\sum_{a \in \mathcal{A}_{\text{valid}}} \pi_\theta(a|s_t)\left(\alpha \log \pi_\theta(a|s_t) - \min_i Q_{\theta,i}(s_t, a)\right)\right]$
    \IF{MCTS distribution available ($m_t = 1$)}
        \STATE $\mathcal{L}_{\text{CE}} \leftarrow -\mathbb{E}_{s_t}\!\left[\sum_a \pi^{\text{MCTS}}(a|s_t) \log \pi_\theta(a|s_t)\right]$
    \ELSE
        \STATE $\mathcal{L}_{\text{CE}} \leftarrow 0$
    \ENDIF
    \STATE $\mathcal{L}_{\text{total}} \leftarrow \mathcal{L}_Q + \mathcal{L}_\pi + \lambda_{\text{mcts}}\, \mathcal{L}_{\text{CE}}$
    \STATE Update $\theta$ by minimizing $\mathcal{L}_{\text{total}}$
    \STATE $\bar{\theta} \leftarrow (1 - \tau)\bar{\theta} + \tau\theta$
\ENDFOR
\end{algorithmic}
\end{algorithm}

\section{PPO+MCTS Training Plots}

\begin{figure}[h]
    \centering
    \begin{subfigure}[h]{0.32\textwidth}
        \centering
        \includegraphics[width=\linewidth]{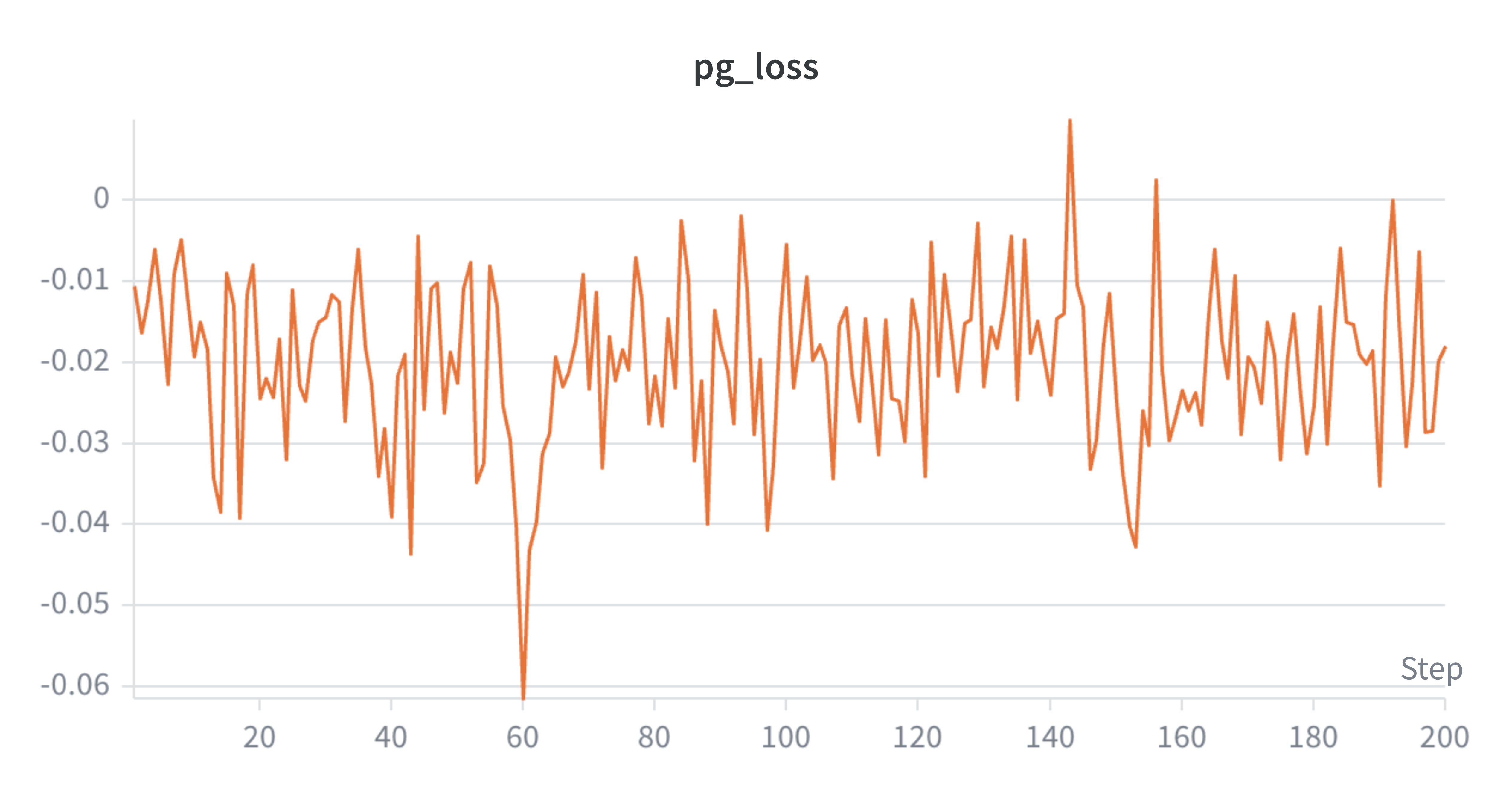}
        \caption{Policy gradient loss}
    \end{subfigure}
    \hfill
    \begin{subfigure}[h]{0.32\textwidth}
        \centering
        \includegraphics[width=\linewidth]{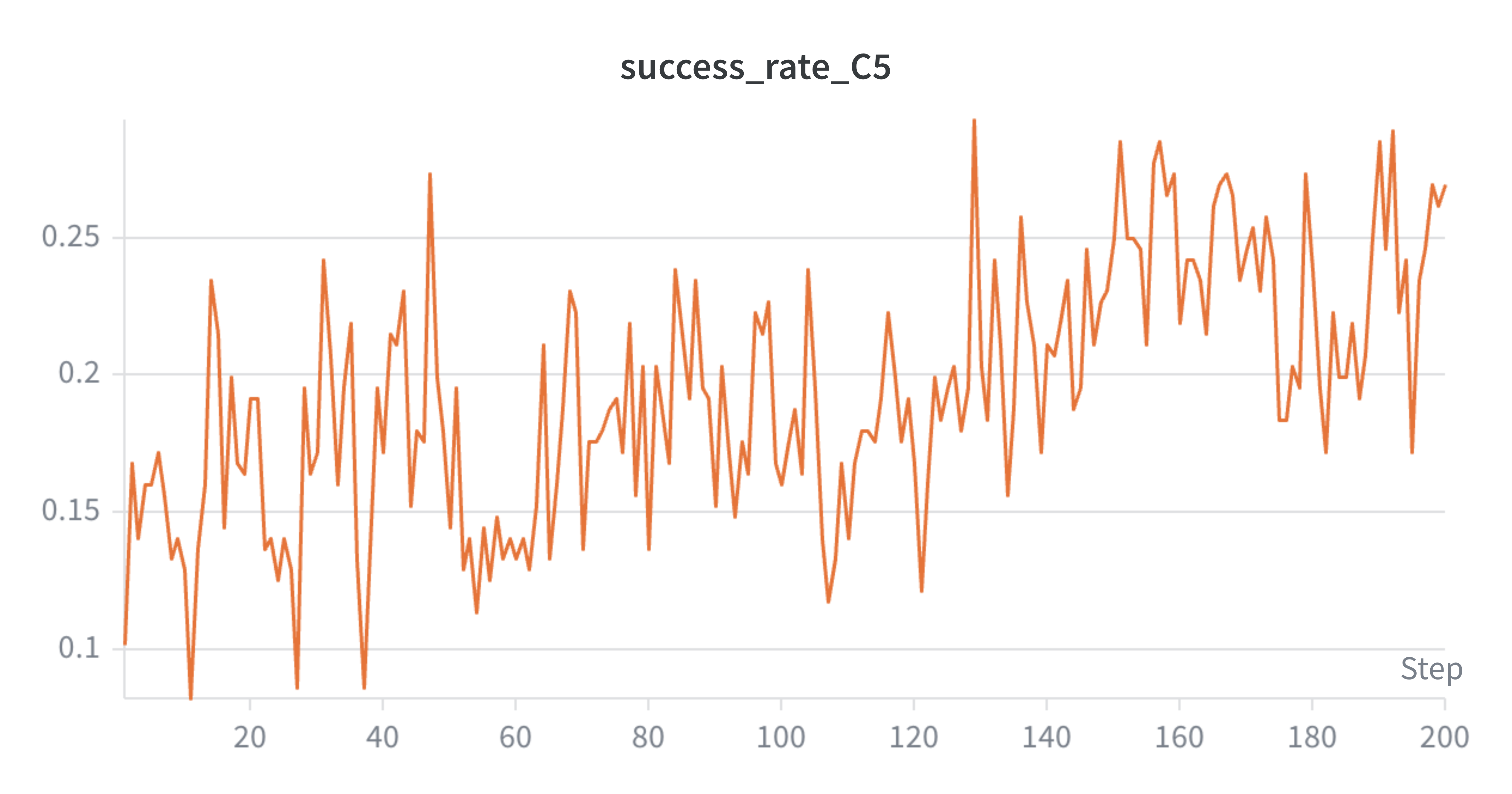}
        \caption{Success rate}
    \end{subfigure}
    \hfill
    \begin{subfigure}[h]{0.32\textwidth}
        \centering
        \includegraphics[width=\linewidth]{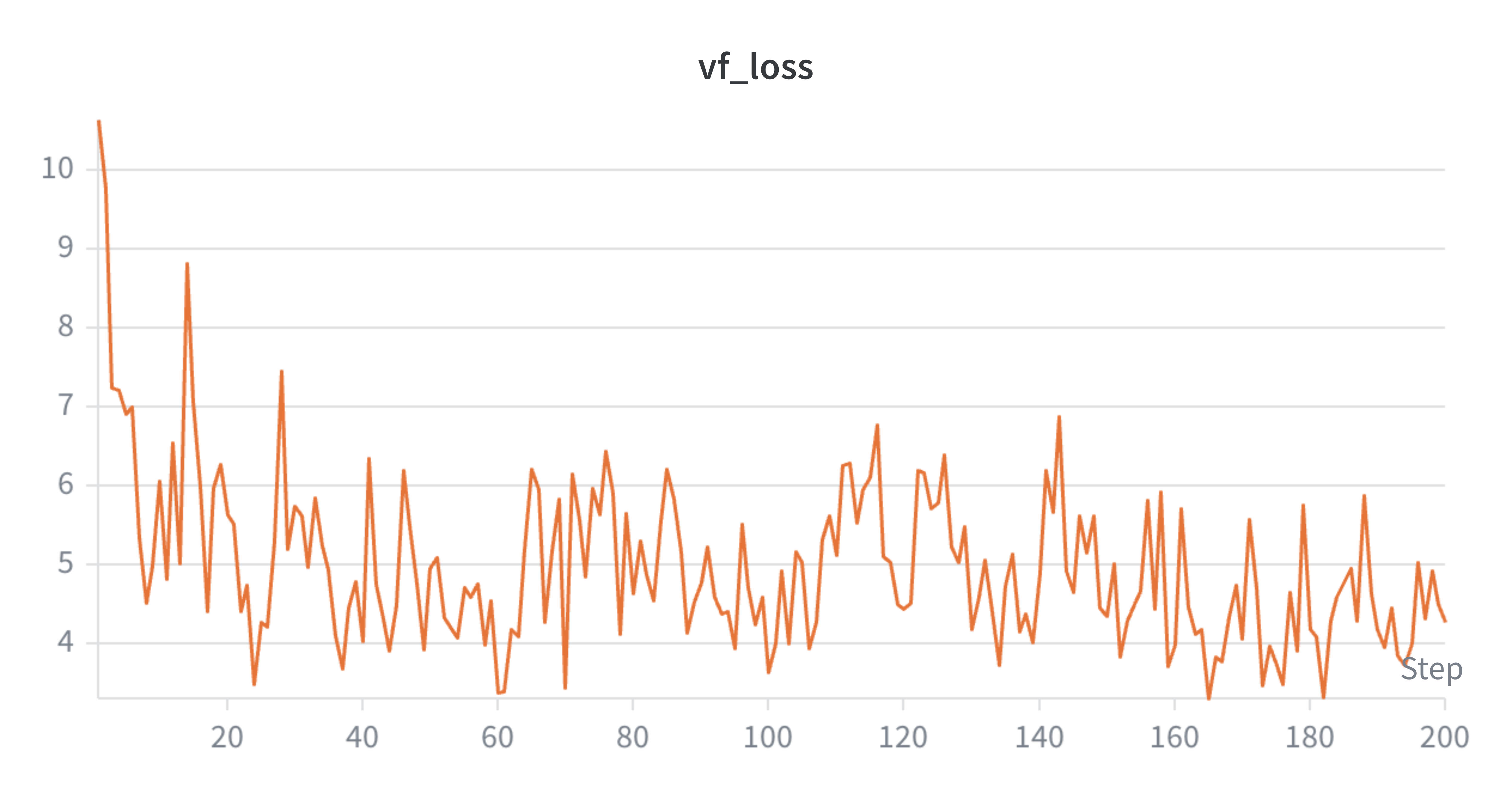}
        \caption{Value function loss}
    \end{subfigure}
    \caption{PPO+MCTS training plots for complexity 5.}
    \label{fig:ppo_mcts_c5_training}
\end{figure}

\begin{figure}[h]
    \centering
    \begin{subfigure}[h]{0.32\textwidth}
        \centering
        \includegraphics[width=\linewidth]{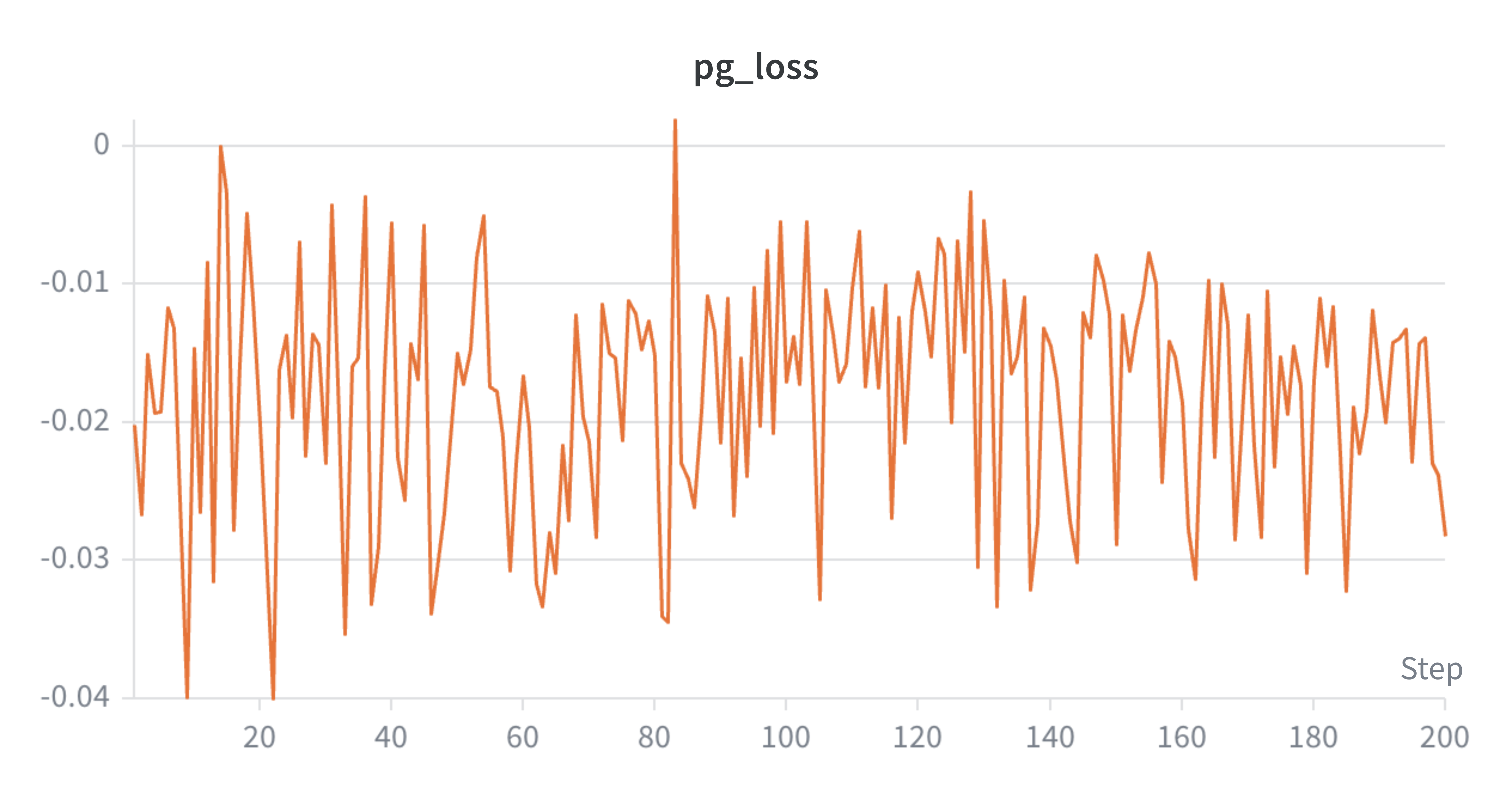}
        \caption{Policy gradient loss}
    \end{subfigure}
    \hfill
    \begin{subfigure}[h]{0.32\textwidth}
        \centering
        \includegraphics[width=\linewidth]{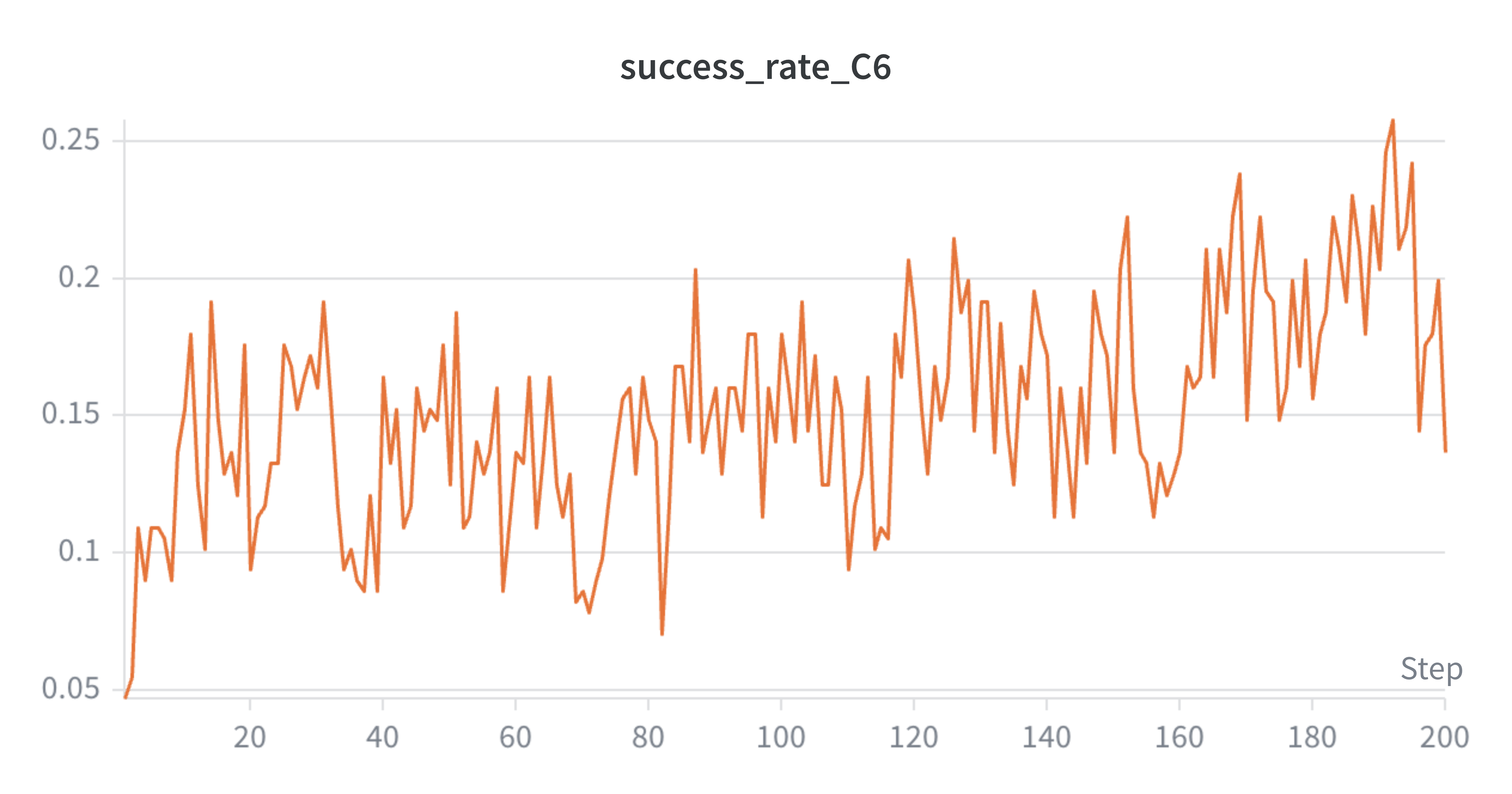}
        \caption{Success rate}
    \end{subfigure}
    \hfill
    \begin{subfigure}[h]{0.32\textwidth}
        \centering
        \includegraphics[width=\linewidth]{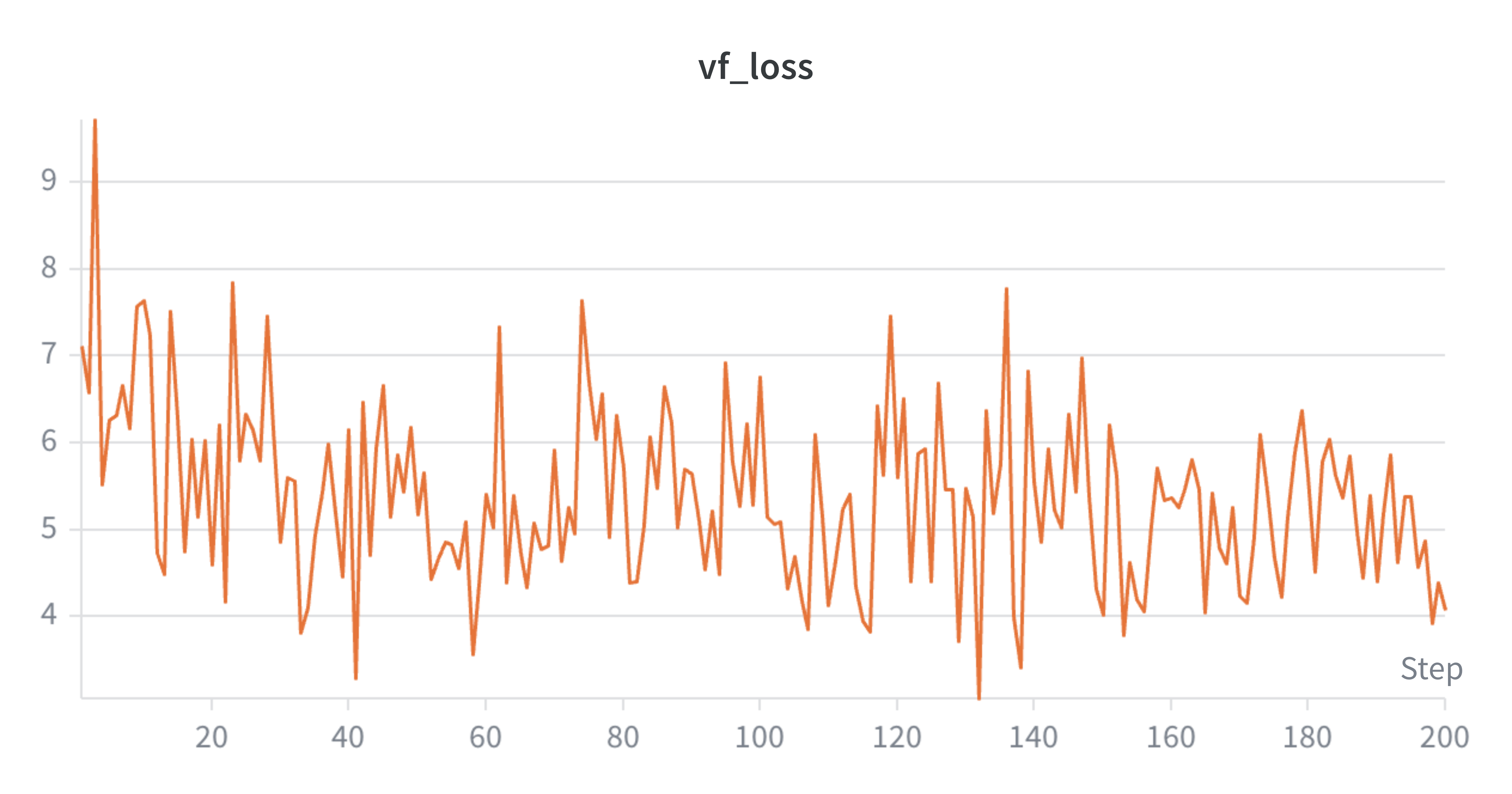}
        \caption{Value function loss}
    \end{subfigure}
    \caption{PPO+MCTS training plots for complexity 6.}
    \label{fig:ppo_mcts_c6_training}
\end{figure}

\newpage
\section{SAC Training Plots}

 \begin{figure}[h]
     \centering
    \includegraphics[width=.9\linewidth]{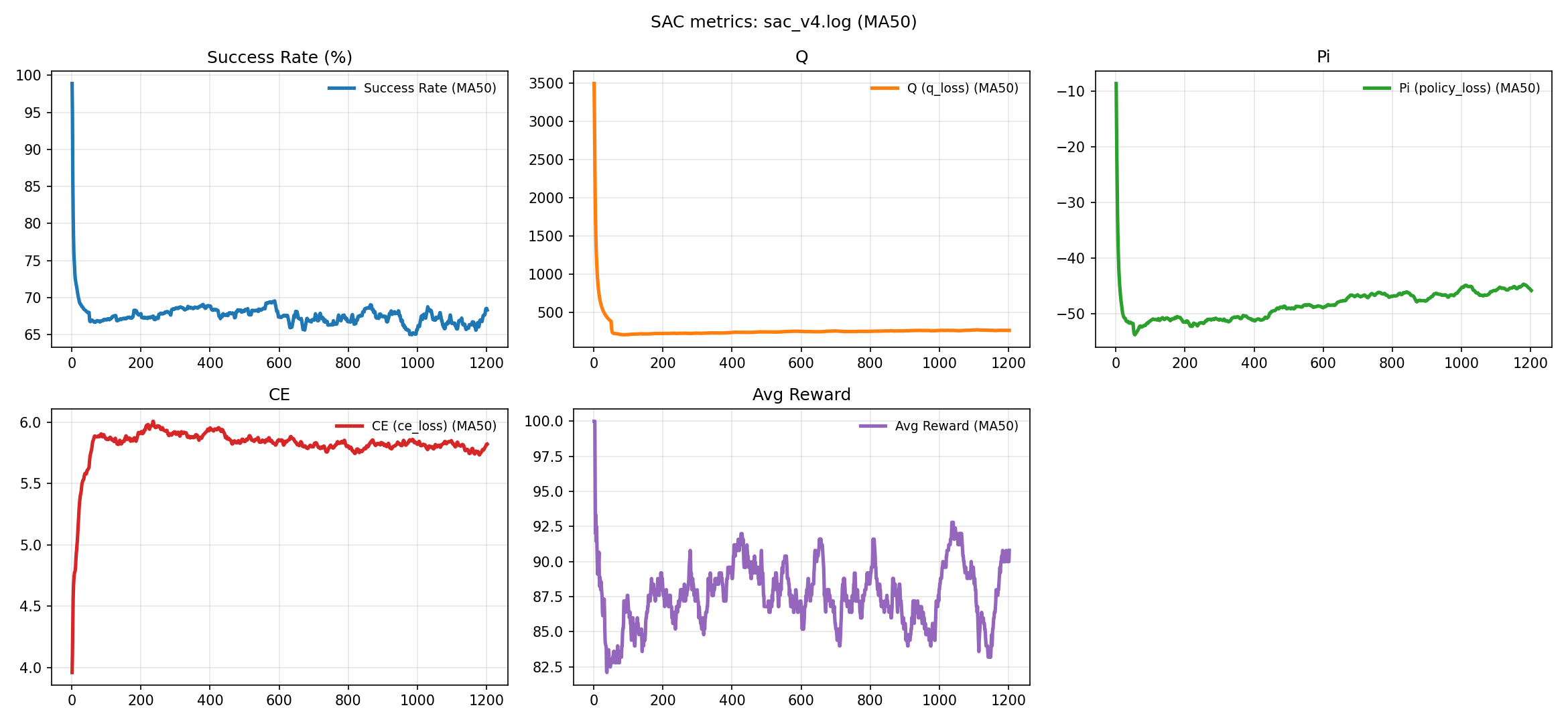}
   \caption{Soft Actor-Critic Training Metrics Over Time (MA50) at Complexity level 1-4}
   \label{fig:SAC}
\end{figure}

 \begin{figure}[h]
     \centering
    \includegraphics[width=.9\linewidth]{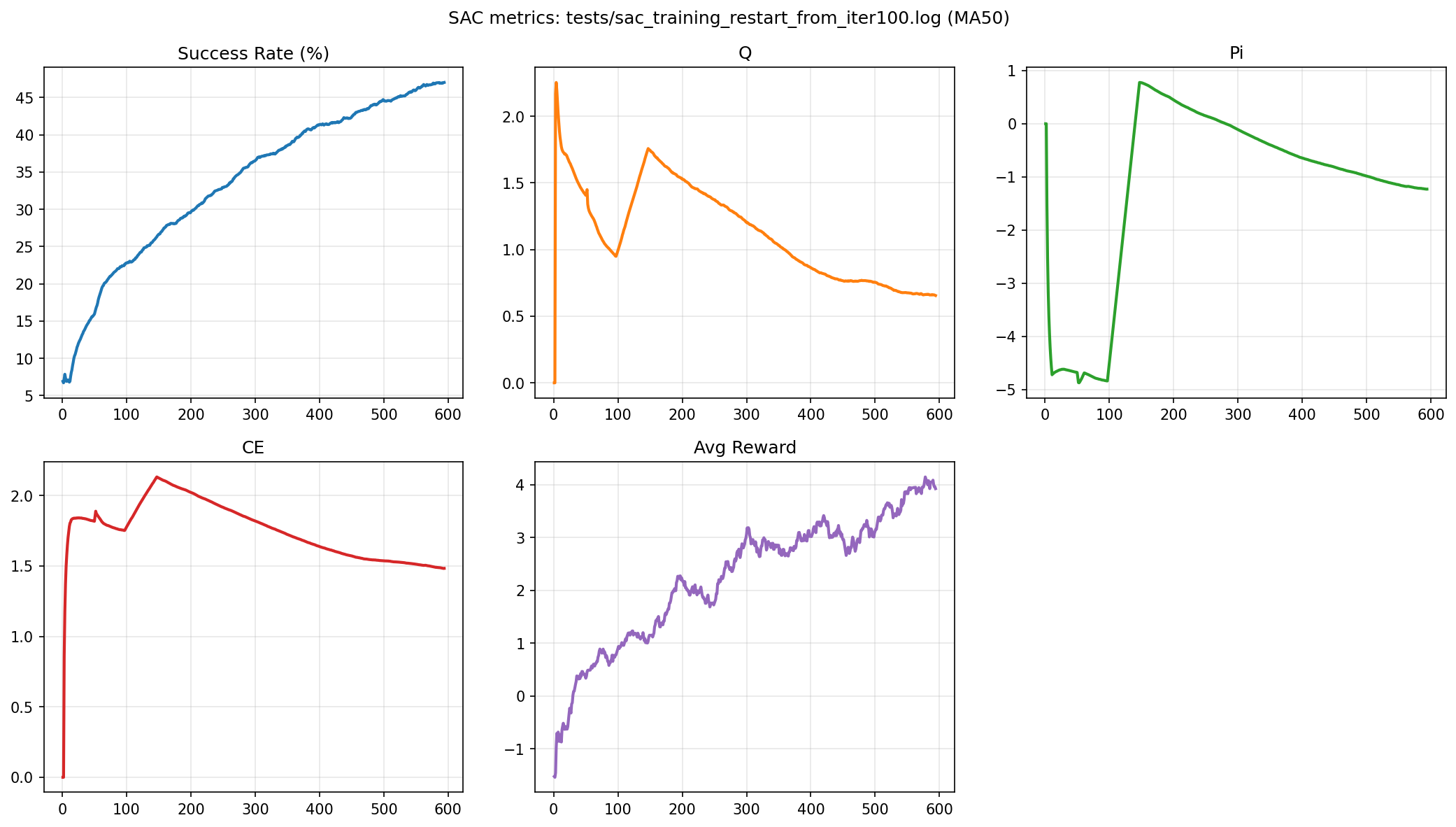}
   \caption{Soft Actor-Critic Training Metrics Over Time (MA50) at Complexity 5}
   \label{fig:SAC-}
\end{figure}

\end{document}